\documentclass{article}
\usepackage{stfloats}
\usepackage{amssymb}
\usepackage{multirow} 
\usepackage{booktabs} 
\usepackage{graphicx} 

\usepackage{spconf,amsmath,graphicx,hyperref}

\usepackage{setspace}

\title{AG-FUSION: ADAPTIVE GATED MULTIMODAL FUSION FOR 3D OBJECT DETECTION IN COMPLEX SCENES}
\name{Sixian Liu$^{1}$ \qquad Chen Xu$^{1}$ \qquad Qiang Wang$^{1 \star}$\thanks{$^\star$ Corresponding author. Email: wangq@bupt.edu.cn} \qquad Donghai Shi$^{2}$ \qquad Yiwen Li$^{1}$}

\address{$^{1}$ National Engineering Research Center for Mobile Network Technologies,\\
Beijing University of Posts and Telecommunications, Beijing 100876, China\\
$^{2}$ Yaowu Technology Co., Ltd, Shenzhen, China\\
}

\begin{document}
%
\maketitle
\begin{abstract}
Multimodal camera-LiDAR fusion technology has found extensive application in 3D object detection, demonstrating encouraging performance. However, existing methods exhibit significant performance degradation in challenging scenarios characterized by sensor degradation or environmental disturbances. We propose a novel Adaptive Gated Fusion (AG-Fusion) approach that selectively integrates cross-modal knowledge by identifying reliable patterns for robust detection in complex scenes. Specifically, we first project features from each modality into a unified BEV space and enhance them using a window-based attention mechanism. Subsequently, an adaptive gated fusion module based on cross-modal attention is designed to integrate these features into reliable BEV representations robust to challenging environments. Furthermore, we construct a new dataset named Excavator3D (E3D) focusing on challenging excavator operation scenarios to benchmark performance in complex conditions. Our method not only achieves competitive performance on the standard KITTI dataset with 93.92\% accuracy, but also significantly outperforms the baseline by 24.88\% on the challenging E3D dataset, demonstrating superior robustness to unreliable modal information in complex industrial scenes.
\end{abstract}
\section{Introduction}
\label{sec:intro}
In recent years, 3D object detection has achieved significant progress on autonomous driving benchmarks \cite{10637966, 10093116, 8794195}. Traditional LiDAR-based methods \cite{9157234,voxel} exploit accurate depth and geometry for strong results, but the sparsity of point clouds limits long-range context and hampers performance. Other approaches projected point clouds into the image space for modality-level feature alignment and synchronous fusion \cite{9878415,9156790,9578812,9423268}, although this strategy often led to a loss of geometric consistency and degradation of semantic information. Subsequent approaches adopted unified Bird's-Eye-View (BEV) representations, which provide geometrically consistent fusion while preserving semantic density and structural integrity \cite{ZHAO2024125103, bevfusion, 9879824}. However, current BEV fusion techniques \cite{bevfusion, li2023bevdepth} mainly rely on convolutional operations, which limit them to static local feature combination and prevent explicit adaptive modeling of cross-modal interactions. Consequently, these methods are not effective in complex environments. 

This paper focuses on autonomous excavator operation scenarios, where perception systems face severe challenges due to complex conditions. Dust and lighting cause significant image degradation, resulting in noisy or blurred visual data. Meanwhile, cluttered backgrounds and articulated parts of machinery lead to frequent occlusions, causing substantial spatial distortion with blurring of the depth \cite{li2023bevdepth}. Currently, while LiDAR provides precise geometric priors, it suffers from point-cloud sparsity \cite{huang2020epnet} and multiple reflection interference on metallic surfaces. These domain-specific challenges severely limit the transferability of autonomous driving-oriented fusion models to industrial environments, making it an open problem to effectively integrate reliable multimodal information under such conditions. 

To address these challenges, we propose the Adaptive Gated Cross-Attention Fusion (AG-Fusion) framework for multimodal 3D object detection, based on BEVFusion \cite{bevfusion}. This approach applies window-based self-attention within each modality to enhance local contextual information. Subsequently, a bidirectional cross-attention module enables explicit interaction between LiDAR and camera features. Finally, a content-adaptive gated mechanism adaptively balances modality contributions, allowing the fusion process to handle occlusion and sensor-specific noise. For evaluation, we construct the Excavator3D dataset (E3D). Experiments on the KITTI \cite{kitti} and E3D data sets demonstrate that AG-Fusion achieves industry-leading detection accuracy and exceptional robustness in automotive and industrial scenarios.

The main contributions of this work are:
1) Adaptive Gated Fusion (AG-Fusion): We propose a novel multimodal 3D object detection architecture. It integrates bidirectional cross-attention with a spatially adaptive gated mechanism on top of enhanced feature extraction.
2) Industrial Excavator Dataset (E3D): We introduce a new multimodal dataset targeting real-world excavator operation, and evaluate our method on both KITTI and E3D, demonstrating strong generalization in complex industrial scenarios.
\section{Method}
\label{sec:format}
\begin{figure*}[t] 
    \centering
    \includegraphics[width=\textwidth]{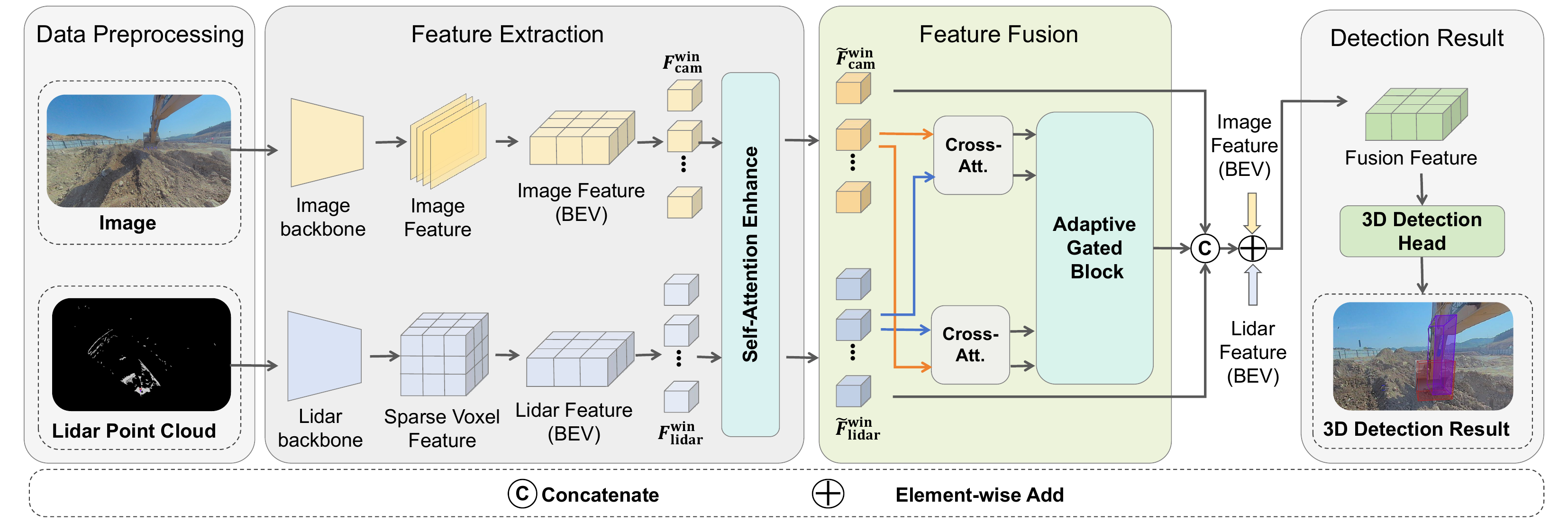} %
    \vspace{-0.7cm}
    \caption{Overview Of The Proposed AG-Fusion Architecture.} %
    \label{fig:1}
    \vspace{-0.5cm}
\end{figure*}

\subsection{Enhanced Feature Extraction}
In industrial excavation scenes, camera-based BEV features often suffer from depth ambiguity and occlusion, while LiDAR-based BEV features are affected by sparsity and reflection noise on metallic surfaces. Inspired by Swin-T \cite{9710580}, we introduce a \textbf{Window-based Self-Attention Enhancement (SA-E)module } that adaptively refines each modality before inter-modal fusion, as shown in Fig. \ref{fig:1}.  

Given a BEV feature map $\textbf{F}_\text{m} \in \mathbb{R}^{H \times W \times C}$ for modality $\text{m} \in \{\text{cam}, \text{lidar}\}$, we partition it into non-overlapping square windows $\{F^\text{win}_\text{m}\}$ of size $h \times w$. Within each window, multi-head self-attention (MSA) is applied to model local feature dependencies:
\vspace{-0.3cm}
\begin{gather}
\tilde{\textbf{F}}^{\text{win}}_\text{m} = \text{MSA}\big( LN(\textbf{F}^\text{win}_\text{m}) \big) + \textbf{F}^\text{win}_\text{m},\\
\hat{\textbf{F}}^{\text{win}}_\text{m} = \tilde{\textbf{F}}^\text{win}_\text{m} + \text{FFN}\big( LN(\tilde{\textbf{F}}^\text{win}_\text{m}) \big),
\end{gather}
where $LN(\cdot)$ denotes layer normalization and FFN denotes a feed-forward network. The enhanced window features $\hat{\textbf{F}}^\text{win}_\text{m}$ are kept in window form and propagated to the subsequent cross-attention fusion module.  

Compared to traditional self-attention with complexity $\mathcal{O}((HW)^2)$, this design reduces the computational cost to $\mathcal{O}(N_\text{win} \cdot (hw)^2)$, where $N_\text{win} = HW / (hw)$ is the number of windows. This efficiency allows the model to process high-resolution BEV features under real-time constraints.  

\subsection{Inter-Modal Cross-Attention Gated Module}
Unlike conventional fusion strategies that perform static or locally constrained feature aggregation, we propose a \textbf{Cross-Attention Gated (CAG)} module. It explicitly models cross-modal interactions through bidirectional attention and adaptively merges features using a data-dependent gated mechanism, making it suitable for challenging industrial environments, such as excavator operation scenarios.

\noindent\textbf{Bidirectional Cross-Attention.}  
Given the enhanced window features $\hat{F}^\text{win}_\text{cam}$ and $\hat{F}^\text{win}_\text{lidar}$, we perform bidirectional cross-attention within each corresponding window region:  
\vspace{-0.05cm}
\begin{gather}
\textbf{A}_{\text{cam} \leftarrow \text{lidar}} = \text{MHA}(q = \hat{\textbf{F}}^\text{win}_\text{cam}, k/v = \hat{\textbf{F}}^\text{win}_\text{lidar}),\\
\textbf{A}_{\text{lidar} \leftarrow \text{cam}} = \text{MHA}(q = \hat{\textbf{F}}^\text{win}_\text{lidar}, k/v = \hat{\textbf{F}}^\text{win}_\text{cam}),
\end{gather}
where $\text{MHA}(\cdot)$ denotes multi-head cross-attention. In this way, the camera stream queries geometric priors from LiDAR features, while the LiDAR stream retrieves semantic and textural cues from the camera.  

\noindent\textbf{Adaptive Gated Fusion.}  
\begin{figure}[t] 
\centering
\includegraphics[width=\linewidth]{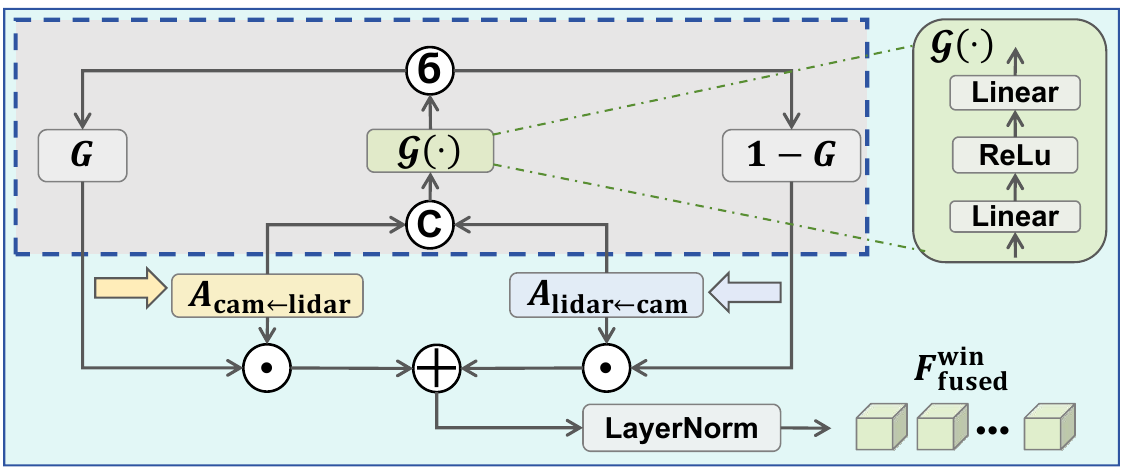} %
\vspace{-0.7cm}
\caption{ Structure of adaptive gated block.} %
\vspace{-0.5cm}
\label{fig:2}
\end{figure}
To integrate the two complementary cross-modal views, we introduce a spatially adaptive gated mechanism, as shown in Fig. \ref{fig:2}. A lightweight sub-network $\mathcal{G}(\cdot)$ generates a pixel-wise gate map:
\begin{equation}
    \textbf{G} = \sigma(\mathcal{G}(\text{Concat}(\textbf{A}_{\text{cam} \leftarrow \text{lidar}}, \textbf{A}_{\text{lidar} \leftarrow \text{cam}}))),
\end{equation}
where $\sigma(\cdot)$ is the sigmoid function. The fused feature is:
\begin{equation}
\textbf{F}^\text{win}_\text{fused} = \textbf{G} \odot \textbf{A}_{\text{cam} \leftarrow \text{lidar}} + (1-\textbf{G}) \odot \textbf{A}_{\text{lidar} \leftarrow \text{cam}},
\end{equation}
with $\odot$ denoting element-wise multiplication. The gated operation adaptively adjusts the modality contributions according to local scene characteristics. For instance, in cases of severe occlusion or LiDAR signal dropout (e.g., around the articulated bucket), the gate can rely more on semantically reliable camera features; in regions where visual ambiguity dominates (e.g., textureless areas or under harsh lighting), it can emphasize geometrically accurate LiDAR features.

The CAG module provides adaptive and context-aware fusion, effectively overcoming the limitations of static convolutional fusion methods such as BEVFusion\cite{bevfusion}. The fused window features $F^\text{win}_\text{fused}$ are subsequently aggregated in the next stage to produce the final multimodal BEV feature.
\vspace{-0.2cm}
\subsection{Multi-Level Feature Aggregation}
After enhanced feature extraction and inter-modal fusion, it is necessary to integrate all feature streams into a unified BEV representation for reliable 3D detection. Formally, the aggregated feature is constructed by channel-wise concatenation followed by a lightweight convolution:
\begin{gather}
\textbf{F}_\text{agg} = \text{Concat}(\hat{\textbf{F}}^\text{win}_\text{cam}, \hat{\textbf{F}}^\text{win}_\text{lidar}, \textbf{F}^\text{win}_\text{fused}),\\
\textbf{F}_\text{out} = \Phi_\text{fuse}(\textbf{F}_\text{agg}),
\end{gather}
where $\Phi_\text{fuse}$ consists of a $1\times1$ convolution, Batch Normalization, and ReLU activation. To stabilize training and preserve important modality-specific cues, we further apply a residual connection with the original BEV features:
\begin{equation}
\textbf{Y} = \text{ReLU}(\textbf{F}_\text{out} + \textbf{F}_\text{cam} + \textbf{F}_\text{lidar}),
\end{equation}
where $\textbf{F}_\text{cam}$ and $\textbf{F}_\text{lidar}$ denote the initial BEV features before enhancement. This residual design facilitates gradient flow while ensuring that the raw modality information is not lost during fusion.   The final aggregated feature $\textbf{Y}$ provides a comprehensive representation for the 3D detection head.
\vspace{-0.3cm}
\subsection{Excavator3D (E3D) Dataset}
\begin{figure}[t] 
\centering
\includegraphics[width=\linewidth]{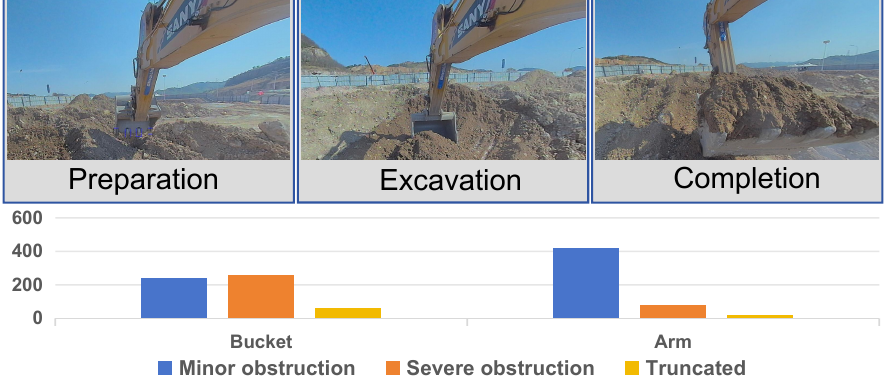} %
\vspace{-0.6cm}
\caption{Example scenes and distribution of minor/severe occlusion and truncation in the E3D dataset.} %
\vspace{-0.5cm}
\label{fig:data}
\end{figure}
The Excavator3D (E3D) dataset was collected from real-world excavator operations under intensive working conditions. It includes synchronized data from a wide-angle LiDAR (0.1--150 m range, 120$^\circ \times$70$^\circ$ FOV, 0.15$^\circ \times$0.36$^\circ$ angular resolution, 192 channels at 905 nm) and an RGB camera (1920$\times$1080 at 22 fps). The dataset encompasses various excavator operation scenarios, with a focus on detecting two key articulated components of the end-effector: the arm and the bucket. As shown in Fig. \ref{fig:data}, the E3D dataset covers three stages of excavator operation: preparation, excavation, and completion. Throughout these stages, frequent self-occlusion and environmental obstructions pose significant challenges for detection. The dataset comprises 500 multi-modal samples with synchronized LiDAR and camera data, each annotated with 3D bounding boxes for both the arm and bucket. The E3D dataset provides a compact yet challenging benchmark for industrial perception research.

\section{experimental evaluation}
\label{sec:pagestyle}
\begin{figure}[t] 
\centering
\includegraphics[width=\linewidth]{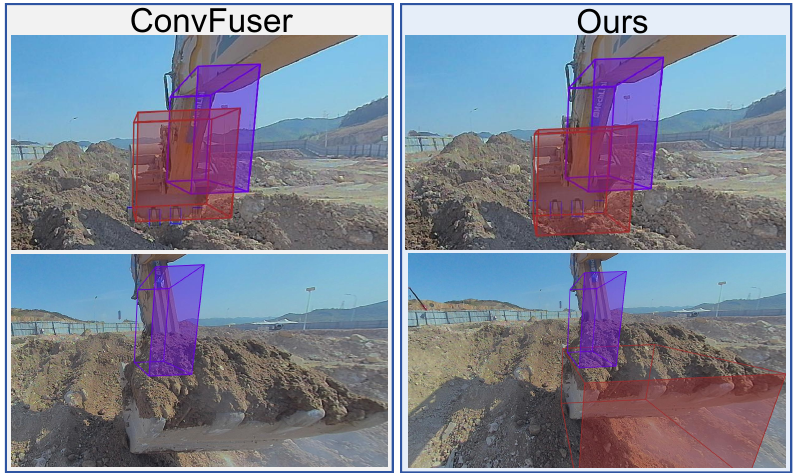} %
\vspace{-0.6cm}
\caption{Performance comparison between the proposed fusion method and BEVFusion on the E3D dataset.} %
\vspace{-0.5cm}
\label{fig:3}
\end{figure}

\subsection{Dataset and Implementation Details}
We built upon the MMDetection3D library \cite{MMDetection3D} and implemented our adaptive fusion framework on the BEVFusion project. Swin-T \cite{9710580} was used as the image backbone and VoxelNet \cite{8578570} as the LiDAR backbone. For KITTI experiments \cite{kitti}, images were resized to $384 \times 1280$ with 1/8 feature resolution in the camera branch, and the voxel size was set to $(0.05, 0.05, 0.1)$ m. The detection range was [0, 70.4] m in $x$, [-40, 40] m in $y$, and [-3, -1] m in $z$. Training was performed using AdamW with cosine annealing (initial learning rate = 0.001), a batch size of 2, and 30 epochs. All experiments were performed on an NVIDIA RTX 4090 GPU.

\vspace{-0.3cm}
\subsection{Main Results}
To comprehensively evaluate the effectiveness of our method, we compare it against state-of-the-art multimodal fusion approaches in the KITTI dataset \cite{kitti}. As shown in Table \ref{tab:1}, our method outperforms all previous leading methods in most metrics. Specifically, on the validation set, it achieves improvements of +1.35\% mAP for the Car class and +2.53\% mAP for the Pedestrian class over the baseline \cite{bevfusion}.

Notably, our method demonstrates substantial superiority in the most challenging scenarios, surpassing the current latest SOTA method, FGU3R \cite{10889148}, by significant margins of +2.42\% and +3.26\% on the 3D Moderate and Hard difficulty levels, respectively. These levels contain the most demanding samples characterized by severe occlusion or extreme distances. This strongly validates that our intra-modality enhancement and adaptive gated fusion mechanism can effectively integrate contextual information over large receptive fields, which is crucial for the precise localization of small and visually ambiguous objects.

As shown in Fig. \ref{fig:3}, the static ConvFuser often does not recognize the excavator bucket under severe occlusion, leading to missing or inaccurate boxes. In contrast, AG-Fusion adaptively balances LiDAR and camera features, maintaining accurate localization in challenging scenes. This comparison highlights the effectiveness of our gated mechanism and supports the quantitative gains in Table \ref{tab:1}.
\begin{table*}[t] 
\centering
\resizebox{\textwidth}{!}
{ 
\setlength{\tabcolsep}{10pt}
\begin{tabular}{@{}cccccccccccc@{}}
\toprule

\multirow{2}{*}{Method} & \multirow{2}{*}{Reference} & \multirow{2}{*}{Modality} & \multicolumn{4}{c}{Car AP$_{\text{3D}}$\%} & \multicolumn{4}{c}{Pedestrian AP$_{\text{3D}}$\%} \\
\cmidrule(lr){4-7} \cmidrule(lr){8-11}
 & & & Easy & Mod. & Hard & mAP & Easy & Mod. & Hard & mAP \\
 
\midrule
PV-RCNN \cite{9157234} & CVPR 2020 & L &92.57 & 84.83 & 82.69& 86.70 & 64.26 & 56.67 & 51.91 & 57.61 \\
Voxel R-CNN \cite{voxel} & AAAI 2021 & L & 92.38 & 85.29 & 82.86& 86.84 & 65.38 & 58.87 & 53.13 & 59.13 \\

\midrule
EPN++ \cite{9983516} & TPAMI 2022 & LC & 92.51 & 83.71 & 81.98& 86.07 & 73.77 & 65.42 & 59.13 & 66.11 \\
CAT-Det \cite{9880265} & CVPR 2022 & LC & 90.12 & 81.46 & 79.15& 83.58 & 54.26 & 45.44 & 41.94 & 47.21 \\
LoGoNet \cite{10205491} & CVPR 2023 & LC & 92.04 & 85.04 & 84.31& 87.13 & 70.20 & 63.72 & 59.46 & 64.46\\
VirConv-T \cite{10205191} & CVPR 2023 & LC & \textbf{95.61} & 87.98 & 86.64& \textbf{90.07} & 73.06 & 66.15 & 59.50 & 66.24 \\
TED-M \cite{wu2023transformation} & AAAI 2023 & LC & 95.55 & 86.48 & 84.26& 88.76 & 72.69 & 65.02 & 58.29 & 65.33 \\
BEVFusion \cite{bevfusion} & ICRA 2023 & LC & 92.85 & 86.98 & 85.33& 88.38 & 73.66 & 67.84 & 62.44 & 67.98\\
FGU3R \cite{10889148} & ICASSP 2025 & LC & 95.26 & 85.84 & 83.67& 88.26 & - & - & - & - \\
Ours & - & LC & 93.92 & \textbf{88.26} & \textbf{86.93} & 89.73 & \textbf{74.51} & \textbf{70.18} & \textbf{66.84} & \textbf{70.51} \\

\bottomrule
\end{tabular}
}
\vspace{-0.3cm}
\caption{Performance Comparison with State-of-the-Art Methods on KITTI val Set for Car and Pedestrian Categories. "Mod." and "-" mean moderate and not mentioned, respectively. Best results are shown in bold.}
\vspace{-0.4cm}
\label{tab:1}
\end{table*}

\vspace{-0.3cm}
\subsection{Ablation Studies}
To systematically validate the effectiveness of the core modules we proposed, we conducted comprehensive ablation studies on both our self-constructed Excavator3D (E3D) industrial scene dataset and the mainstream KITTI dataset \cite{kitti}. 
\begin{table}[h!]
\renewcommand{\arraystretch}{0.7}
\vspace{-0.5cm}
\centering
\resizebox{\columnwidth}{!}{
\begin{tabular}{cc|ccc|ccc}
\toprule
\multicolumn{2}{c}{Component} & \multicolumn{3}{c}{Car AP$_{\text{3D}}$\%} & \multicolumn{3}{c}{Car AP$_{\text{BEV}}$\%} \\
\cmidrule(lr){1-2}\cmidrule(lr){3-5} \cmidrule(lr){6-8}
SA-E & CAG & Easy & Mod. & Hard & Easy & Mod. & Hard\\
\midrule
 &  & 92.85 & 86.98 & 85.33 & 93.05 & 88.92 &86.73\\[1ex]
\checkmark &  & 93.63 & 86.93 & 84.45 & 93.85 & 89.77 & \textbf{88.65}\\[1ex]
 & \checkmark &93.28& 87.43& 84.84& 93.59&89.39& 87.54\\[1ex]
\checkmark & \checkmark & \textbf{93.92}&\textbf{88.26}&\textbf{86.93}&\textbf{93.91}&\textbf{90.13}&88.61\\[1ex]
\bottomrule
\end{tabular}
}
\vspace{-0.4cm}
\caption{Ablation study on the KITTI dataset.}
\vspace{-0.3cm}
\label{table:kitti_ablation}
\end{table}

\begin{table}[h!]
\renewcommand{\arraystretch}{0.6}
\vspace{-0.2cm}
\centering
\setlength{\tabcolsep}{3pt}
\resizebox{\columnwidth}{!}{
\begin{tabular}{c|ccc|ccc}
\toprule
\multirow{2}{*}{Fusion}& \multicolumn{3}{c}{Bucket} & \multicolumn{3}{c}{Arm} \\
\cmidrule(lr){2-4} \cmidrule(lr){5-7}
 & AP$_{\text{BEV}}$\% & P\% & R\% & AP$_{\text{BEV}}$\% & P\% & R\%\\
\midrule
ConvFuser \cite{bevfusion} &52.62& 53.76&31.31& 98.58&96.75& 95.33\\[1ex]
Fixed G=0.3& 67.54 & 68.29 & 46.06 & 97.50 & 97.56 &95.20\\[1ex]
Fixed G=0.7 & 61.09 & 62.04 & 39.82 & 98.64 & 97.31 &96.21\\[1ex]
AdaptiveGate&\textbf{77.50}$\uparrow$&\textbf{78.08}$\uparrow$&\textbf{59.90}$\uparrow$&97.04&97.11&95.29\\[1ex]
\bottomrule
\end{tabular}
}
\caption{Performance comparison of different fusion strategies on the E3D dataset. }
\vspace{-0.4cm}
\label{table:e3d_ablation}
\end{table}

\noindent\textbf{Component-wise Contribution Analysis.} We performed module ablation studies on KITTI to evaluate each component's contribution, as shown in Table \ref{table:kitti_ablation}. When employed individually, the SA-E module demonstrated consistent improvements across most metrics, achieving notable gains of 0.8\%, 0.85\%, and 1.92\% on the BEV Easy, Moderate, and Hard levels, respectively. The CAG module also provided appreciable improvement, validating the superiority of its cross-modal fusion strategy. The synergistic combination of both modules yielded the best overall performance across nearly all difficulty levels and evaluation metrics, achieving state-of-the-art results of 88.26\% and 86.93\% on the particularly revealing 3D Moderate and Hard benchmarks, which are most indicative of model robustness.

\noindent\textbf{Effectiveness of the CAG Module.} We evaluated the performance of different fusion strategies on the E3D dataset, as shown in Table \ref{table:e3d_ablation}. Compared to the ConvFuser used in BEVFusion \cite{bevfusion}, even a simple fixed-weight gated strategy provided a significant performance boost, underscoring the critical importance of the fusion strategy in industrial scenarios. Our proposed CAG Module markedly outperformed all baseline methods, elevating the $\text{AP}_{\text{BEV}}$ for the most challenging Bucket category from a baseline of 52.62\% to 77.50\%, an absolute improvement of 24.88\%.
\vspace{-0.3cm}
\section{Conclusion}

This paper proposes a multimodal 3D detection framework featuring a Cross-Attention Gated (CAG) module. To overcome the limitations of static fusion under occlusion and sensor noise, our approach employs window-based self-attention for enriched feature extraction, along with a content-adaptive gating mechanism that dynamically integrates LiDAR and camera features. Evaluated on KITTI and the newly introduced E3D dataset, the method achieves notable improvements of 1.35\% mAP on cars and 2.53\% on pedestrians compared to BEVFusion. Significant gains on moderate and hard cases demonstrate enhanced robustness in complex environments. Future work will focus on dataset expansion and optimizing the fusion module for real-time applications.

\label{sec:typestyle}



\vfill\pagebreak

\label{sec:refs}

\begin{small}
\section{Acknowledgements}
This paper was partially funded by the National Natural Science Foundation of China under Grant 62071066, the National Key R\& D Program of China under Grant 2020YFB1806602 and BUPT-Yaowutech cooperation.
\bibliographystyle{ieeetr}
\bibliography{main}
\end{small}
\end{document}